\documentclass[conference]{IEEEtran}
\IEEEoverridecommandlockouts
\usepackage{microtype}
\usepackage{cite}
\usepackage{amsmath,amssymb,amsfonts}
\usepackage{algorithmic}
\usepackage{graphicx}
\usepackage{textcomp}
\usepackage{xcolor}
\usepackage{hyperref}
\usepackage{amsmath}
\def\BibTeX{{\rm B\kern-.05em{\sc i\kern-.025em b}\kern-.08em
    T\kern-.1667em\lower.7ex\hbox{E}\kern-.125emX}}
\begin{document}

\title{Short-Term Electricity Demand Forecasting of Dhaka City Using CNN with Stacked BiLSTM\\
}

\author{
\IEEEauthorblockN{
Kazi Fuad Bin Akhter\IEEEauthorrefmark{1}, 
Sadia Mobasshira\IEEEauthorrefmark{2}, 
Saief Nowaz Haque\IEEEauthorrefmark{3}, 
Mahjub Alam Khan Hesham\IEEEauthorrefmark{4}, 
Tanvir Ahmed\IEEEauthorrefmark{5}
}
\IEEEauthorblockA{\IEEEauthorrefmark{1}Tennessee State University, Nashville, TN\\
Email: fuadbinakhter@gmail.com}
\IEEEauthorblockA{\IEEEauthorrefmark{2}\IEEEauthorrefmark{3}\IEEEauthorrefmark{4}\IEEEauthorrefmark{5}Ahsanullah University of Science and Technology, Dhaka, Bangladesh\\
Emails: sadiamobashira@gmail.com, saiefpulock@gmail.com, mahjub98@gmail.com, tanvir.cse@aust.edu}
}

\maketitle

\begin{abstract}
The precise forecasting of electricity demand also referred to as load forecasting, is essential for both planning and managing a power system. It is crucial for many tasks,
including choosing which power units to commit to, making plans for future power
generation capacity, enhancing the power network, and controlling electricity
consumption. As Bangladesh is a developing country, the electricity infrastructure is
critical for economic growth and employment in this country. Accurate forecasting of
electricity demand is crucial for ensuring that this country has a reliable and sustainable
electricity supply to meet the needs of its growing population and economy. The complex
and nonlinear behavior of such energy systems inhibits the creation of precise algorithms.
Within this context, this paper aims to propose a hybrid model of Convolutional Neural Network
(CNN) and stacked Bidirectional Long-short Term Memory (BiLSTM) architecture to perform an accurate
short-term forecast of the electricity demand of Dhaka city. Short-term forecasting is
ordinarily done to anticipate load for the following few hours to a few weeks.
Normalization techniques have been also investigated because of the sensitivity of these
models towards the input range. The proposed approach produced the
best prediction results in comparison to the other benchmark models (LSTM, CNN-BiLSTM and CNN-LSTM) used in the study, with MAPE 1.64\%, MSE 0.015, RMSE 0.122 and MAE 0.092. The result of the proposed model also outperformed some of the existing works on load-forecasting.
\end{abstract}

\begin{IEEEkeywords}
Deep Learning, Short-term forecasting,  
 \mbox{Electricity} Demand Forecasting, Neural Networks
\end{IEEEkeywords}

\section{Introduction}
Electricity demand forecasting is the process of estimating future electricity consumption for a particular region or system. It is an important tool for utilities, power grid operators, and energy market participants, as it allows them to plan for and manage electricity supply and demand in an efficient and cost-effective way. Electricity is a flexible form of energy, a vital resource for modern life, and a critical input for economic development. Electricity is always in high demand in all economies, particularly among households and businesses. In Bangladesh, the industrial as well as agricultural sectors heavily rely upon electricity as a source of energy. Between 2000 and 2021, Bangladesh's electricity usage rose by 526\%\cite{ghc}. Moreover, Bangladesh's per-capita electricity consumption has gone up significantly and is projected to increase by 22 times by 2050 compared to 2014.  \cite{ghc}. 

The foundation of energy investment planning is precise demand forecasting. It is crucial to the operation and decision-making of the energy market. Underestimating energy usage leads to power outages and malfunctions throughout the electrical grid, whilst overestimating it wastes financial resources. Even just a 1\% increase in error rate for load forecasting can end up in a \$10 million increase in annual operational expenses. \cite{bunn}. Therefore, short-term demand forecasting is getting popular among researchers due to its growing importance in power systems, particularly in smart-grids and micro-grids \cite{sadie}.

Neural networks, a well-accepted machine learning method, have been effectively used to solve a variety of real-world problems, including speech recognition, video analysis, image classification, weather forecasting, stock market prediction, and energy consumption prediction \cite{rodrigo}. Recently, deep neural networks have attracted the research community's interest due to their capacity to capture patterns in data while dealing with enormous volumes of data. As neural networks can investigate not only classification but also regression tasks, they can be used directly as regressors for electricity demand forecasts.
In this paper, we demonstrated the practical implications of deep neural networks to forecast the daily electricity demand of Dhaka city. We have proposed  Convolutional Neural Network (CNN) with stacked
Bidirectional Long-Short Term Memory (BiLSTM) architecture for forecasting. This study is conducted on real-world data consisting of
daily consumption of electricity in Dhaka for the past 6 years (2016-2021). We additionally observed the models that have been employed in prior research.

Our work is summarized in the following points:
\begin{itemize}
    \item The experiment is conducted on real-world data consisting of daily consumption of electricity in Dhaka for the past 6 years (2016-2021). All these data have been collected from the official government websites of Bangladesh.
    \item A CNN with stacked BiLSTM model has been utilized to forecast the electricity demand on a short-term basis.
    \item To compare the effectiveness of the models, we evaluated them using different performance metrics.
\end{itemize}

The following is how the paper is structured. Section II describes in detail the different methods and approaches used to forecast energy consumption in a range of research and experiments published in the literature. Section III then describes some studies regarding deep learning models used in this paper. Section IV then demonstrates how to put the proposed technique into action. Following that, Section V covers simulation results and analysis, and Section VI concludes the article.

\section{Related Works}

Several studies have been performed to predict electricity demand based on different input data. Some of these works are highlighted here about the proposed work. In \cite{son}, long-short-term memory (LSTM),  autoregressive integrated moving average (ARIMA), and multiple linear regression (MLR) were used to train and test the collected data, which consisted of 22 years worth of monthly data (264 observations) on electricity demand in South Korea. Social and weather-related characteristics were also taken into account in this study. The LSTM architecture was able to attain the best predicting score with a Mean Absolute Percentage Error (MAPE) value, of 0.07. 

The authors in \cite{Emre}, used 5 months of hourly electricity consumption data. The goal of this study was the estimation of the next-hour demand for a microgrid. As inputs, the current date, time, and electricity consumption were used, and the demand for the following hour was considered as output. The authors applied an artificial neural network (ANN) with backpropagation to predict the hourly consumption of the area. The prediction error was 11.7412\% according to Root Mean Squared Error (RMSE) Percentage. 

In \cite{bibi}, The authors acquired previous information on hourly power usage in Panama from 2016 to 2019. Several machine learning and deep learning models were compared and benchmarked in this study to predict Panama's short-term electricity consumption. Support Vector Regression (SVR), XGBoost, AdaBoost, Random Forest, and LightGBM were used as machine learning, and deep learning regression, Bidirectional LSTM (Bi-LSTM), and GRU were among the deep learning methods used as part of this study. With a MAPE of 2.90\%, the best-estimating demand of one hour in advance was achieved by deep learning regression.

The study in \cite{ali}, looked at a variety of methods for forecasting electricity demand on Gokceada Island in Turkey, including ANN, Particle Swarm Optimization (PSO), and MLR. The goal of this study was to determine which socio-economic variables have the greatest influence on managing the growth of electricity demand. The dataset consisted of monthly electricity demand data for the period 2014–2019. In this case, the ANN model performed best, with an overall R regression value of 0.99773.

In \cite{miona}, The electricity usage dataset was collected from a facility for cold storage that uses a consumption measuring process that provides data every hour. The obtained data cover the years 2019 to 2021, comprising a total of 18,960 data points gathered over two years. The models used were ordinary RNN, LSTM, GRU, bidirectional LSTM, and bidirectional GRU. The best result was obtained with GRU with a MAPE value of 2.54\% for the spring season and LSTM with a MAPE value of 1.615\% for the winter season.

The proposed study of the authors in \cite{sousa} was utilized with information from the London Households dataset "Smart Meter Energy Consumption Data," which included half-hourly measurements from a sample of 5567 houses. The goal was to forecast the load profile for the upcoming day using only previous load values and a few auxiliary variables. The applied models were Multivariate Adaptive Regression Splines, ANN as well as Random Forests (RF). When compared to the baseline, ANNs were shown to be the most accurate model when compared to the baseline, with a 15\% reduction in Mean Absolute Error. 

The author of the study \cite{majed}, applied deep neural network (DNN), ANN also decision tree for short-term load forecasting. Hourly electricity demand data (200 observations) of the Ontario province in Canada along with weather data were used in this study. According to the findings, the DNN model outperformed the other models with the highest \( R^{2} \) score and was statistically distinct from them.

\section{Background Study}

\subsection{Convolutional Neural Network}

 Convolutional neural networks (CNNs) are a form of deep learning neural networks that are useful for both image processing tasks and forecasting issues due to their capacity to learn and extract features from enormous amounts of information. The input layer, convolutional, pooling, fully connected layers and output layer are among the layers that commonly make up a CNN's architecture . The input layer receives data for the analysis. After that, a number of convolutional and pooling layers process the data. By obtaining the maximum or average value over a restricted area, pooling layers downsample the input data. The network becomes more effective and the size of the input data is decreased. The information is then passed through one or more fully linked layers. This enables the network to learn more complex patterns in the data. The output layer generates the  final output e.g. classification label or a forecasted number.

\subsection{Long Short-Term Memory}
Long short-term memory (LSTM) is a type of recurrent neural network (RNN) that is generally used for forecasting time series data, like stock prices or weather data.The input gate $i$, the forget gate $f$, and the output gate $o$ are the three gates with memory cell $c$ that are responsible for the information flow into and out of each LSTM cell. The forward pass formulae for an LSTM cell with a forget gate can be written in the following forms:

\begin{equation}
\begin{aligned}
f_t &= \sigma g\left( W_f x_t + U_f h_{t-1} + b_f\right) \\
i_t &= \sigma g\left( W_i x_t + U_i h_{t-1} + b_i\right) \\
o_t &= \sigma g\left( W_o x_t + U_o h_{t-1} + b_o\right) \\
\bar{c}_t &= \sigma g\left( W_c x_t + U_c h_{t-1} + b_c\right) \\
c_t &= f_t \odot c_{t-1} + i_t \odot \bar{c}_t \\
h_t &= o_t \odot \sigma h \left(c_t \right)
\end{aligned}
\end{equation}

here,$x_t$ input vector, $f_t$ forget gate's activation vector,$i_t$ input gate's activation vector, $o_t$ output gate's activation vector, $h_t$ hidden state vector, $c_t$ cell state vector and $\bar{c}_t$ cell input activation vector. The operator $\odot$ and the $W$ respectively indicate element wise product and the wight matrices. Lastly, $t$ is the time step.

\subsection{Bidirectional Long Short-Term Memory}

Bidirectional Long Short-Term Memory (BiLSTM) is a type of RNN that can process sequential data using both past and future information. It is composed of two separate LSTM networks where one is responsible for processing the input sequence forward (Equation \ref{eq-f-lstm}) and the other of which processes it backward(Equation \ref{eq-b-lstm}.\break

For the forward LSTM :

\begin{equation}
\begin{aligned}
f_t^{(f)} &= \sigma(W_f^{(f)} x_t + U_f^{(f)} h_{t-1}^{(f)} + b_f^{(f)}) \\
i_t^{(f)} &= \sigma(W_i^{(f)} x_t + U_i^{(f)} h_{t-1}^{(f)} + b_i^{(f)}) \\
o_t^{(f)} &= \sigma(W_o^{(f)} x_t + U_o^{(f)} h_{t-1}^{(f)} + b_o^{(f)}) \\
\bar{c}_t^{(f)} &= \tanh(W_c^{(f)} x_t + U_c^{(f)} h_{t-1}^{(f)} + b_c^{(f)}) \\
c_t^{(f)} &= f_t^{(f)} \odot c_{t-1}^{(f)} + i_t^{(f)} \odot \bar{c}_t^{(f)} \\
h_t^{(f)} &= o_t^{(f)} \odot \tanh(c_t^{(f)})
\end{aligned}
\label{eq-f-lstm}
\end{equation}

Here, $x_t$ is input at time $t$. $f_t^{(f)}$ is forget gate, $i_t^{(f)}$ is input gate, $o_t^{(f)}$ output gate and $\bar{c}_t^{(f)}$ is candidate cell state. $c_t^{(f)}$ and $h_t^{(f)}$ are cell state and hidden state at time $t$. $h_{t-1}^{(f)}$ is hidden state at time $t - 1$. $W$,$U$ are weight matrices and $b$ is bias vector for forward LSTM.\break

For the backward LSTM :

\begin{equation}
\begin{aligned}
f_t^{(b)} &= \sigma(W_f^{(b)} x_t + U_f^{(b)} h_{t+1}^{(b)} + b_f^{(b)}) \\
i_t^{(b)} &= \sigma(W_i^{(b)} x_t + U_i^{(b)} h_{t+1}^{(b)} + b_i^{(b)}) \\
o_t^{(b)} &= \sigma(W_o^{(b)} x_t + U_o^{(b)} h_{t+1}^{(b)} + b_o^{(b)}) \\
\bar{c}_t^{(b)} &= \tanh(W_c^{(b)} x_t + U_c^{(b)} h_{t+1}^{(b)} + b_c^{(b)}) \\
c_t^{(b)} &= f_t^{(b)} \odot c_{t+1}^{(b)} + i_t^{(b)} \odot \bar{c}_t^{(b)} \\
h_t^{(b)} &= o_t^{(b)} \odot \tanh(c_t^{(b)})
\end{aligned}
\label{eq-b-lstm}
\end{equation}

Here, for backward LSTM all the terms are the same as forward LSTM, where $(f)$ is replaced by $(b)$.\break

A single output sequence is created by combining the output of both LSTMs. A sequence of the previous information is then fed into the network in order to train the BiLSTM to predict the following value in the sequence. The input layer collects the input sequence and turns it into a sequence of vectors so that the hidden layers may process it. Each of the BiLSTM cells in the hidden layers processes a different segment of the input sequence. Each BiLSTM cell creates a new hidden state as well as an output vector after receiving as input a vector from the input sequence and the previous hidden state.

\section{Methodology}
\begin{figure*}[t]
\centerline{\includegraphics[width=\textwidth]{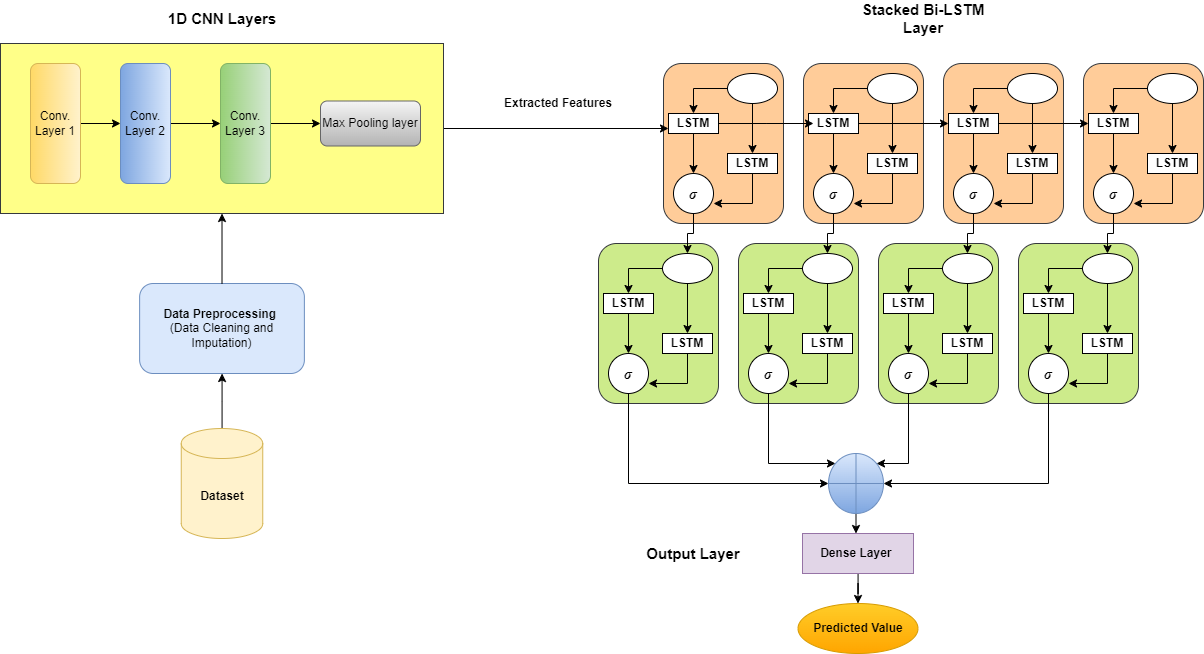}}
\caption{Proposed Methodolgy.}
\label{fig}
\end{figure*}

This section describes the procedures followed in this study. The entire methodology is depicted in Fig.\ref{fig}.

\subsection{Dataset Collection}
For our study, we have used daily electricity consumption data of Dhaka, the capital city of Bangladesh. All of the data was collected from the Bangladesh Power Development Board\cite{bpdb} and the Power Grid Company of Bangladesh\cite{pgcb}. Data spanning 6 years (2016-2021, a total of 2190 observations) were applied for training and testing. The pattern of \mbox{electricity} consumption in Dhaka city varies across weekdays, as it is \mbox{ influenced} by various factors such as industrial, \mbox{commercial} and \mbox{residential} activities. It also depends on weather \mbox{conditions}, and \mbox{population} density of the city. Electricity demand is generally higher during the weekdays (Sunday-Thursday), when commercial and industrial activity is at its peak. However, demand decreases on weekends (Friday and Saturday) as these activities remains inactive. The average electricity demand on weekdays in that period (2016-2021) was about 3300MW, whereas in weekends it stayed between 2900MW to 3250MW.  The dataset includes a timestamp and zone-wise electricity usage for each day. For load forecasting only Dhaka zone's electricity usage data is used.

\subsection{Data Preprocessing}
The dataset contains daily electricity consumption data from 01-01-2016 to 31-12-2021. Due to the manual generation of reports, there were some missing data as well as some incorrect data. The presence of missing or incorrect data in the context of the electricity consumption dataset could have a significant impact on the accuracy and effectiveness of any insights or models derived from the data. So, first we identify the missing data and remove incorrect ones. Then, to impute the missing data we have used the linear interpolation\cite{xu2022} method. Linear interpolation is used in data imputation to estimate missing values in a sequence of integers. 
\begin{equation}
y= y1 + \frac{(x-x_{1})(y_{2}-y_{1})}{(x_{2}-x_{1})}
\label{eq0}
\end{equation}
In the equation \ref{eq0}, the known data points are $(x1,y1)$ and $(x2,y2)$, x is the place where the value is missing (to be imputed), and y is the imputed value.
 The input consists of zone-wise electricity demand (in this paper we used Dhaka zones) at the sub-station end.

\subsection{Splitting Input Data}
For this paper, we have split the dataset into two separate sets - train and test. Training set was used to train the models and a test set was used to evaluate the models. We have split the dataset into 80:20 ratio , where 80\% was used for training and 20\% for testing in order to evaluate the model results.

\subsection{Normalization}
Normalization is the process of transforming the values of numeral columns in a dataset to a similar scale without information loss or distortion of variations in value ranges. We have used Min max scaling as normalization technique. Min-max normalization technique is a linear transformation of the original data. Variables measured at different scales
do not contribute similarly to the model fitting, learned function and can also result in bias.
To address this potential issue, feature-wise normalization, such as MinMax Scaling, is typically used prior to model fitting. For each feature, all values are transformed to a decimal between 0 and 1, with 0 as the least value and 1 as the greatest. The process is based on the \textcolor{blue}{Equation}\ref{ref:eq1} :
\begin{equation}
 X_{scaled} = \frac{x-x_{min}}{x_{max}-x_{min}}
\label{ref:eq1}
\end{equation}

Where, $x$ is a collection of the observed values in $X$, $x_{min}$ is the set of smallest values in $X$ and  $x_{max}$ is the set of largest values in $X$. 

\subsection{Proposed Model}

Convolutional neural networks (CNNs) combined with a stack of bidirectional long short-term memory (BiLSTM) networks have been used in this study for load forecasting. The CNN part of the model is responsible for extracting features from the input data, while the BiLSTM parts are responsible for modeling the temporal dependencies in the data. The output of the CNN is typically fed into the input of the BiLSTM, which processes the data and produces a forecast for the next time step.

The input data is fed into the CNN, which consists of three convolutional layers. Each convolutional layer is made up of a set of filters which are typically small and are applied to the input data to extract features acording to equation \ref{eq 0.1}. Where, $x_{i,j}=$ CNN output at the i-th location and the j-th filter, $b_{j}=$ bias term for the j-th filter, $w_{m,j}=$ weight of the m-th input for the j-th filter, $x_{i+m}=$ m-th input at the i-th position, $M=$ kernal size.

\begin{equation}
    x_{i,j}=b_{j}+\sum_{m=0}^{M-1}w_{m,j}\cdot x_{i+m}
    \label{eq 0.1}
\end{equation}

Each convolutional layer's output is sent via a nonlinear activation function, ReLU (rectified linear unit) as described in equation\ref{eq 0.2}. Here, $x_{i,j}=$ CNN layers' output and $f({x_{i,j}})=$ output of ReLU activation function.

\begin{equation}
    f({x_{i,j}})=max(0,x_{i,j})
    \label{eq 0.2}
\end{equation}

The ouput of equation\ref{eq 0.2} passes through a max pooling layer, which downsamples the data by taking the maximum feature values within a small neighborhood. In equation\ref{eq 0.3}, $y_{i,j}=$ max-pooling layer at the i-th position and j-th filter, $f(x_{k,j})=$ ReLU function output at the k-th position and j-th filter and  $P=$ pooling window size.

\begin{equation}
      y_{i,j}=max(f(x_{k,j})),\forall k \in [i,i+P-1]
      \label{eq 0.3}
\end{equation}

The output of the CNN (in this case, $y_{i,j}$) is a set of extracted features, which are passed to the BiLSTM. The BiLSTM consists of a sequence of LSTM cells, each of which takes as input the extracted features from the CNN and the output from the previous LSTM cell. The LSTM cells are connected in a bidirectional manner, meaning that they process the data in both forward and backward directions.

The equations after passing $y_{i,j}$ to the forward LSTM are ,

\begin{equation}
\begin{aligned}
f_{t}^{(f)} &= \sigma(W_{f}^{(f)} \cdot [h_{t-1}^{(f)}, y_{i,j}] + b_{f}^{(f)}) \\
i_{t}^{(f)} &= \sigma(W_{i}^{(f)} \cdot [h_{t-1}^{(f)}, y_{i,j}] + b_{i}^{(f)})  \\
\bar{C}_{t}^{(f)} &= \tanh(W_{C}^{(f)} \cdot [h_{t-1}^{(f)}, y_{i,j}] + b_{C}^{(f)}) \\
C_{t}^{(f)} &= f_{t}^{(f)} * C_{t-1}^{(f)} + i_{t}^{(f)} * \bar{C}_{t}^{(f)} \\
o_{t}^{(f)} &= \sigma(W_{o}^{(f)} \cdot [h_{t-1}^{(f)}, y_{i,j}] + b_{o}^{(f)}) \\
h_{t}^{(f)} &= o_{t}^{(f)} * \tanh(C_{t}^{(f)})
\end{aligned}
\end{equation}

Similarly for backward LSTM,  

\begin{equation}
\begin{aligned}
f_{t}^{(b)} &= \sigma(W_{f}^{(b)} \cdot [h_{t+1}^{(b)}, y_{i,j}] + b_{f}^{(b)}) \\
i_{t}^{(b)} &= \sigma(W_{i}^{(b)} \cdot [h_{t+1}^{(b)}, y_{i,j}] + b_{i}^{(b)}) \\
\bar{C}_{t}^{(b)} &= \tanh(W_{C}^{(b)} \cdot [h_{t+1}^{(b)}, y_{i,j}] + b_{C}^{(b)}) \\
C_{t}^{(b)} &= f_{t}^{(b)} * C_{t+1}^{(b)} + i_{t}^{(b)} * \bar{C}_{t}^{(b)} \\
o_{t}^{(b)} &= \sigma(W_{o}^{(b)} \cdot [h_{t+1}^{(b)}, y_{i,j}] + b_{o}^{(b)}) \\
h_{t}^{(b)} &= o_{t}^{(b)} * \tanh(C_{t}^{(b)})
\end{aligned}
\end{equation}

Here, The symbol [,] indicates concatenation. $tanh$ is the hyperbolic tangent activation function, while $\sigma$ is the sigmoid activation function and $*$ is the element-wise multiplication. The forward and backward hidden states are concatenated to generate the BiLSTM layer's final output, $h_{t}$ = [$h_{t}^{(f)}$,$h_{t}^{(b)}$]. The output of the max-pooling layer (output of the CNN) contains the local features extracted from the input data. On the other hand, the hidden state of the BiLSTM contains information about the sequence processed so far, including potential long-term dependencies. With the concatenation of these two types of information, the model can make predictions based on both the local features of the input data and the long-term dependencies in the sequence.

As it is a stacked BiLSTM architecture, the input for the subsequent BiLSTM layer is the output of the previous BiLSTM layer. Each layer has a distinct set of weights and biases, but the equations stay the same. Usually, each layer's initial cell state and hidden state are set to zero.The output then passed through a fully connected (dense) layer,  before being used to make the forecast. 

The model starts with a 1-dimensional CNN layer (Conv1D is a convolutional layer built exclusively
for analyzing one-dimensional data, such as time series) with 64 filters, a 3-dimensional kernel size,'same' padding, and a ReLU activation function. The model then adds two additional CNN layers, each with 128 or 256 filters, a kernel size of 3, the'same' padding, and the ReLU activation function. A max pooling layer with pool size 2 is placed after each of these CNN layers. The model then contains stacked BiLSTM layers after the CNN layers. The first BiLSTM layer contains 256 units and is configured to return sequences, allowing it to send sequence information to the following LSTM layer. he second BiLSTM layer also has 256 units but does not return sequences. Finally, the model includes a dense layer with a single unit.

\begin{table}[htbp]
\centering

\caption{Proposed Model Configuration}
\label{table1}
\resizebox{7cm}{!}{%
\begin{tabular}{|c|c|}
\hline
\textbf{Model Parameters} & \textbf{Values}    \\ \hline
Optimizer                 & Adam               \\ \hline
Loss Function             & Mean Squared Error \\ \hline
Batch Size                & 64                 \\ \hline
Eporchs                   & 500                \\ \hline
\end{tabular}%
}
\end{table}

For model configuration, table\ref{table1}, 'Adam' optimizer is used, 'mean squared error' is used as loss function,
batch size is 64 and the model is run for 500 epochs.

\subsection{Evaluation Metrices}
The primary goal of this study is to reduce error between actual and predicted electricity demand. In this study, we have used four different metrics - Mean Absolute Percentage Error (MAPE), Mean Absolute Error (MAE), Mean Squared Error (MSE) and Root Mean Squared Error (RMSE) to evaluate the model's performance. 

MAPE measures the prediction accuracy of a forecasting model with the equation :
\begin{equation}
MAPE = \frac{1}{n} \sum_{t=1}^{n} \left| \frac{A_t - F_t}{A_t} \right| * 100
\end{equation}
where, $n$ is number of data points, $A_t$ is actual load and $F_t$ is forecasted load at time $t$. MAE determines the average magnitude of the errors in a set of predictions with, 

\begin{equation}
MAE = \frac{1}{n} \sum_{t=1}^{n} |A_t - F_t|
\end{equation}
MSE calculates average squared difference between the forecasted values and the actual value. The equation is :
\begin{equation}
MSE = \frac{1}{n} \sum_{t=1}^{n} (A_t - F_t)^2
\end{equation}
Finally, RMSE is the square root of the MSE.
\begin{equation}
RMSE = \sqrt{\frac{1}{n} \sum_{t=1}^{n} (A_t - F_t)^2}
\end{equation}

\subsection{Experimental Setup}
This study was conducted using Google Colaboratory. Tesla K80 GPU with 12GB memory, 13GB of RAM, and also Intel Xeon CPU. This study also employed the Python version 3.7.1 programming language and libraries, including Keras version 2.8.0, 2.8.2 version of TensorFlow. The dataset was compiled manually from trustworthy government websites.

\section{Results and Analysis}

To evaluate the efficiency of our proposed model, we compared the performance of our proposed approach to some other works regarding forecasting electricity demand in table \ref{table-2}. In comparison to those models, the proposed approach achieves a MAPE value of 1.64\%.

\begin{table}[htbp]
\centering
\caption{Performance comparison of the proposed model with some existing studies}
\label{table-2}
\begin{tabular}{|c|c|c|}
\hline
\textbf{Work}                                 & \textbf{Procedure}       & \textbf{MAPE (\%)} \\ \hline
Son et. al \cite{son}        & LSTM                     & 7                  \\ \hline
Ibrahim et. al \cite{bibi}   & Deep Learning Regression & 2.9                \\ \hline
Stošović et. al \cite{miona} & GRU                      & 2.54               \\ \hline
\textbf{Proposed Model}                       & \textbf{CNN with Stacked BiLSTM} & \textbf{1.64}      \\ \hline
\end{tabular}
\end{table}

The table\ref{table-2} shows that the proposed model has achieved MAPE reduction between 35.28\% to 76.51\% compared to the methods used in the mentioned studies. Moreover, The performance of the proposed model was also evaluated against the performance of four benchmark models (LSTM, CNN-LSTM, CNN-BiLSTM) that were trained during the study. 

\begin{table}[htbp]
\caption{Models' Performance Metrics comparison with benchmark models}
\begin{center}
\begin{tabular}{|c|c|c|c|c|}
\hline
\textbf{Model} & \textbf{MSE} & \textbf{RMSE} & \textbf{MAE} & \textbf{MAPE (\%)}  \\ \hline
 LSTM &  0.0153 & 0.1510 & 0.0943 & 1.6624 \\ \hline
 CNN-BiLSTM & 0.0153 & 0.1236 & 0.0933 & 1.6488  \\ \hline
 CNN-LSTM &  0.0153 & 0.1236 & 0.0936 & 1.6547  \\ \hline
 \textbf{Proposed Method}   &  \textbf{0.0151} & \textbf{0.1229} & \textbf{0.0929} & \textbf{1.6445}\\ \hline
\end{tabular}
\label{table3}
\end{center}
\end{table}

Table\ref{table3} shows that the proposed model has better performance than the 3 benchmark models - LSTM, CNN-LSTM, and CNN-BiLSTM. It has the lowest MAPE value which indicates that the proposed model's forecasted results are the closest to the real values, in comparison to the other benchmarks models. In terms of mean absolute error (MAE), mean square error (MSE), and root mean square error (RMSE), the suggested model performs better than the other three models. The mean differences between the predicted and actual data is the least in the model with the lowest MAE. Moreover, it has the lowest MSE, demonstrating the extent to which the model reduces larger errors. The model also shows the lowest root mean square error (RMSE), showing significant data around the line of best fit. These results suggest that the proposed model has a better accuracy rate for predictions.

\section{Conclusion}
The purpose of this study was to propose and develop an accurate forecasting model for daily regional electricity demand. CNN with stacked BiLSTM method gave the best prediction result with MAPE value of 1.6445\%. By precisely estimating daily electricity demand, which is responsible for the largest proportion of overall electricity usage, the suggested model is intended to contribute to effective power-system planning. Further study will focus on considering multiple data sources to improve the performance of the proposed approach, such as weather information, socio-economic factors, and population information. Separating data from public holidays and regular workdays can improve performance and provide more diverse information.

\bibliographystyle{IEEEtran}
\bibliography{References}

\end{document}